%% file: cvpr.tex
\documentclass[final]{cvpr}

\usepackage{times}
\usepackage{epsfig}
\usepackage{graphicx}
\usepackage{amsmath}
\usepackage{amssymb}
\usepackage{enumitem}
\usepackage{multirow}
\usepackage{amsmath}
\DeclareMathOperator{\Tr}{Tr}

\usepackage[pagebackref=true,breaklinks=true,colorlinks,bookmarks=false]{hyperref}

\usepackage{xcolor}
\definecolor{bittersweet}{rgb}{1.0, 0.44, 0.37}
\definecolor{mygreen}{rgb}{0.29, 0.7, 0.48}

\newcommand{\modelName}{MITR}

\makeatletter
\newcommand{\thickhline}{%
    \noalign {\ifnum 0=`}\fi \hrule height 1pt
    \futurelet \reserved@a \@xhline
}
\makeatother

\usepackage{caption} 

\makeatletter
\renewcommand{\paragraph}{%
  \@startsection{paragraph}{4}%
  {\z@}{0.5ex}{-1em}%
  {\normalfont\normalsize\bfseries}%
}
\makeatother

\def\cvprPaperID{3613} 
\def\confYear{CVPR 2021}

\usepackage{mathtools}

\DeclarePairedDelimiter\floor{\lfloor}{\rfloor}
\usepackage{hyperref}

\begin{document}

\title{
Connecting What to Say With Where to Look \\ by Modeling Human Attention Traces
}



\author{
Zihang Meng\textsuperscript{1}, 
Licheng Yu\textsuperscript{2}, 
Ning Zhang\textsuperscript{2}, 
Tamara Berg\textsuperscript{2}, \\
Babak Damavandi\textsuperscript{2},
Vikas Singh\textsuperscript{1}, 
and Amy Bearman\textsuperscript{2}  \\ \\
\textsuperscript{1}University of Wisconsin Madison ~~~~ \textsuperscript{2}Facebook AI \\
{\tt\small zmeng29@wisc.edu, vsingh@biostat.wisc.edu} \\
{\tt \small \{lichengyu,ningzhang,tlberg,babakd,abearman\}@fb.com}
}


\maketitle

\thispagestyle{empty}

\input{text/0.abstract}
\input{text/1.intro}

\input{text/2.related_work}
\input{text/3.method}

\input{text/4.experiments}
\input{text/5.conclusion}

\clearpage 
\clearpage
{\small
\bibliographystyle{ieee_fullname}
\bibliography{cvpr}
}

\clearpage

\appendix

\input{Localized Narratives/supp.tex}

\end{document}

%% file: text/0.abstract.tex
\begin{abstract}
We introduce a unified framework to jointly model images, text, and human attention traces.
Our work is built on top of the recent Localized Narratives annotation framework~\cite{pont2019connecting}, where each word of a given caption is paired with a mouse trace segment. 
We propose two novel tasks: (1) predict a trace given an image and caption (i.e., visual grounding), and (2) predict a caption and a trace given only an image.
Learning the grounding of each word is challenging, due to noise in the human-provided traces and the presence of words that cannot be meaningfully visually grounded.
We present a novel model architecture that is jointly trained on dual tasks (controlled trace generation and controlled caption generation).
To evaluate the quality of the generated traces, we propose a local bipartite matching (LBM) distance metric which allows the comparison of two traces of different lengths.
Extensive experiments show our model is robust to the imperfect training data and outperforms the baselines by a clear margin.
Moreover, we demonstrate that our model pre-trained on the proposed tasks can be also beneficial to the downstream task of COCO's guided image captioning. Our code\footnote{Code: \href{https://github.com/facebookresearch/connect-caption-and-trace}{github.com/facebookresearch/connect-caption-and-trace}} and project page\footnote{Project page: \href{http://pages.cs.wisc.edu/\~zihangm/connect\_caption\_trace}{http://pages.cs.wisc.edu/{\raise.17ex\hbox{$\scriptstyle\sim$}}zihangm/connect\_caption\_trace}} are publicly available.

\end{abstract}

%% file: text/1.intro.tex
\section{Introduction}


\begin{figure}[!ht]
\centering
\includegraphics[width=0.46\textwidth]{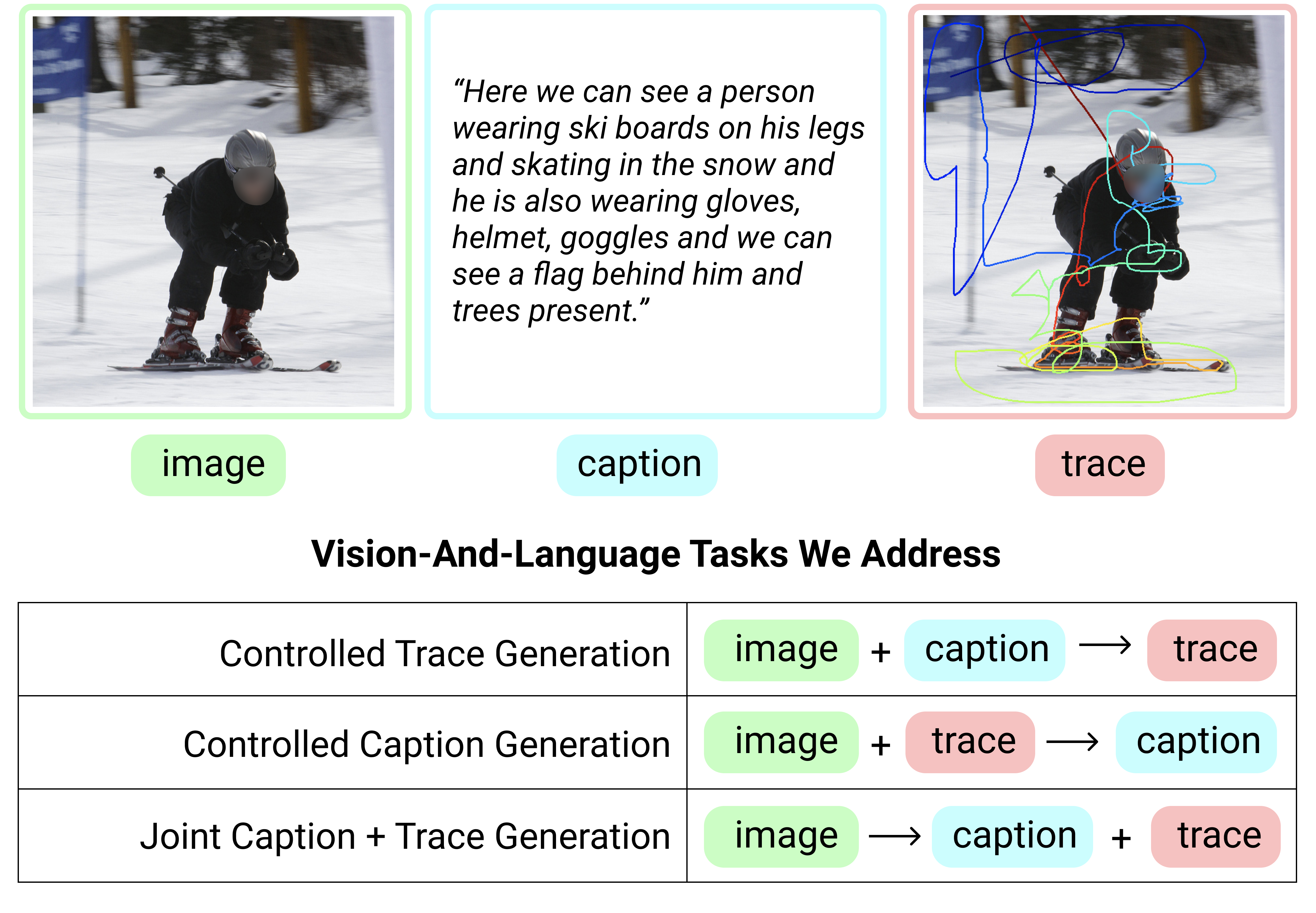} 
\caption{The three vision-and-language tasks, as illustrated on a single example from the Localized Narratives dataset. The first and third depicted tasks are novel.}
\label{fig:three_tasks}
\end{figure}

The development of powerful models and algorithms within computer vision and natural language processing proceeded along distinct trajectories with only occasional overlap until recently. 
However, ideas from these two fields are gradually converging, with a focus on building multi-modal models, particularly for aligning visual and language stimuli~\cite{ViLBERT_NeurIPS_2019,tan2019lxmert,Su2020VLBERT,chen2019uniter}.
The goal of these models is to mimic humans' extraordinary abilities to compress information and translate it across modalities.
Several joint or combined visual recognition and natural language understanding tasks have emerged as natural tests of these vision-and-language models' capabilities.
\emph{Image captioning} asks a model to identify and localize the key scene elements in an image and describe them in natural language form.
\emph{Visual grounding}, and specifically \emph{phrase localization}, asks a model to solve the reverse problem: given a natural language query, identify the target object(s) of the query in the image.
\emph{Controlled image captioning}, first introduced in~\cite{cornia2019show}, combines the two tasks. 
Here, an external user is asked to specify which parts of the image they want described and in what order (e.g., by providing an ordered sequence of bounding boxes). 
The output captions are therefore explicitly grounded in the image. 
One application of this line of work is automatically generating localized descriptions of images for visually impaired users on social media services.
This removes the need to rely on human-written ``alt'' text, which is often missing in web images~\cite{Bigham_2006}. 

Vision-and-language models share common components and techniques.
Image captioning architectures are typically composed of two modules: an image encoder, which ingests and interprets an image, and a language model decoder, which generates a natural language caption~\cite{xu2015showDenseCap,Huang_2019_ICCV}. 
Visual grounding models first identify the key components of the image (i.e., bounding box proposals) and query (i.e., which words or phrases to focus on), extract features from each, and then correlate them to predict the referred-to object~\cite{Rohrbach_2016,Hu_2016a,plummer2015flickr30k,yu2016modeling}.
Architectures for both tasks often rely on attention~\cite{xu2015showDenseCap,Deng_2018_CVPR,Huang_2019_ICCV,ViLBERT_NeurIPS_2019}, a mechanism inspired by the human visual system~\cite{Rensink_2000,Corbetta_2002}.
Researchers have also designed more complex models that can do both caption generation and grounding. 
For example, \cite{Mao_2016_CVPR} and~\cite{yu2016modeling} can both generate an unambiguous description of a specific object or region in an image and automatically select an object given a referring text expression.

Despite these advancements, existing image captioning and visual grounding models cannot jointly generate longform, natural language captions \emph{and} dense, word-level visual groundings.
This is because existing image captioning datasets only provide short captions with sparse groundings at the noun level (Flickr30k Entities~\cite{plummer2015flickr30k}) or phrase level (Google RefEx~\cite{Mao_2016_CVPR}, Flickr30k Entities~\cite{plummer2015flickr30k} and Visual Genome~\cite{krishna2016visualgenome}).
To address these limitations,~\cite{pont2019connecting} introduced the Localized Narratives dataset, in which annotators were asked to describe an image with their voice while simultaneously drawing a mouse trace over the region they are describing.
This annotation framework provides rich, longform image captions and dense visual grounding in the form of a mouse trace segment for each word.
The work in \cite{pont2019connecting} incorporates the annotated mouse trace to aid in standard image captioning and controlled image captioning tasks.
However, it does not investigate the reverse problem of directly predicting the mouse trace or explore the connections between caption generation and trace generation.

In this paper, we take a step beyond~\cite{pont2019connecting} by requiring models to directly predict the trace, which is analogous to a fine-grained and temporally grounded log of human attention. 
Besides \emph{controlled caption generation}, where a model generates a caption guided by the given ordered trace from~\cite{pont2019connecting}, we further introduce two challenging new tasks: \emph{controlled trace generation}, where a model must densely localize each word from a natural language caption in an image, and \emph{joint caption and trace generation}, where a model is only given an image and must act as an annotator in the Localized Narratives protocol.
There tasks are shown in Fig.~\ref{fig:three_tasks}.
{The task of predicting the trace is meaningful in two ways. First, a point-wise trace is a straightforward means for representing eye gaze. Learning the trace (independent of specific use cases) could be variously useful, and this is made possible by the efficient collection scheme described in \cite{pont2019connecting} which does not rely on expensive gaze trackers. Second, this form of annotation yields ``weakly-labeled'' word-level grounding. We demonstrate that such ``weak'' word-to-trace alignment could offer benefits for some important vision and language tasks. Besides, the predicted trace can provide a better explanation than most attention-based image captioning approaches.}
To evaluate the generated traces, we propose a novel evaluation metric, local bipartite matching (LBM), to compare two traces of arbitrary length.
We present a flexible new transformer-based model architecture that is trained in parallel on controlled caption generation and controlled trace generation. 
The model also incorporates a symmetric cycle loss to improve the quality of the generated caption and trace. 
In addition to the three tasks mentioned above, we show that our approach can benefit downstream tasks by pre-training on our proposed tasks before fine-tuning for the downstream setting.



To summarize, we make the following contributions:
\begin{itemize}
    \setlength\itemsep{0em}
    \item  We introduce two novel tasks: ($i$) controlled trace generation and ($ii$) joint caption and trace generation.
    \item We present a novel mirrored transformer model architecture (\modelName), which is jointly trained and evaluated on three vision-and-language tasks.
    \item We design an evaluation metric to address the challenge of computing the distance between two traces of different lengths.
    \item By jointly learning from the mirrored trace generation task, our proposed method benefits the downstream task of guided caption generation on the COCO dataset.
\end{itemize}

%% file: text/2.related_work.tex
\section{Related Work}

\paragraph{Image Captioning} 
Image captioning is typically formulated using a generative model, creating descriptions in textual space given the input image via CNN-to-RNN/LSTM/Transformer~\cite{vinyals2015show, chen2015mind, karpathy2015deep, cornia2020m2}.
An increasingly common addition to this basic architecture is a visual attention
mechanism which typically produces a spatial map that identifies the specific image
region(s) most relevant to the current word prediction task~\cite{xu2015show, anderson2018bottom}.
However, the learned spatial attention may not well align with human attention~\cite{das2017human}. 
To model attention more directly, controlled image captioning was first introduced in \cite{xu2015showDenseCap}.
It requires the user to provide a sequence of bounding boxes in the image and outputs the image caption in the same order, describing the objects in those bounding boxes. 
The authors in \cite{pont2019connecting} adjusted the task by using an annotator's mouse trace for the control. 




\paragraph{Visual Grounding} 
The task of visual grounding is to localize a region described by a given text query.
Researchers have introduced multiple datasets to tackle this problem, such as RefCOCO~\cite{yu2016modeling}, Google RefEx~\cite{Mao_2016_CVPR}, Flickr30K~\cite{plummer2015flickr30k}, and DenseCap~\cite{johnson2016densecap}.
State-of-the-art approaches~\cite{yu2018mattnet, yu2017joint, liu2017referring, zhang2017discriminative} treat visual grounding as selecting the most matched box to the input text query.
However, the input query is typically short (the average length of captions in RefCOCO is 3.5 words) and the grounding is sparse (each query corresponds to just a single box).
By contrast, our work focuses on denser word-to-region grounding.
 

\paragraph{Localized Narratives} 
As described above, image captioning datasets only provide image-sentence pairs without the spatial localization of words. Visual grounding datasets only provide sparse region-sentence mapping. 
Recently, Localized Narratives~\cite{pont2019connecting} was proposed, which offers dense word-region alignment for each full caption.
This dataset was collected by recording annotators' voice and mouse traces simultaneously when describing the image content.
The three modalities of image, trace, and caption significantly expands the scope of how we can connect vision and language.
While~\cite{pont2019connecting} only addressed a single task of controlled captioning, we introduce two more novel and challenging tasks, i.e., controlled trace generation, and joint caption and trace generation.
At first glance, these three tasks as shown in Fig.~\ref{fig:three_tasks} appear separate; however, we propose a unified framework using a mirrored transformer to jointly model all three tasks.








%% file: text/3.method.tex
\section{Method}

\begin{figure*}[!ht]
\centering
\includegraphics[width=0.95\textwidth]{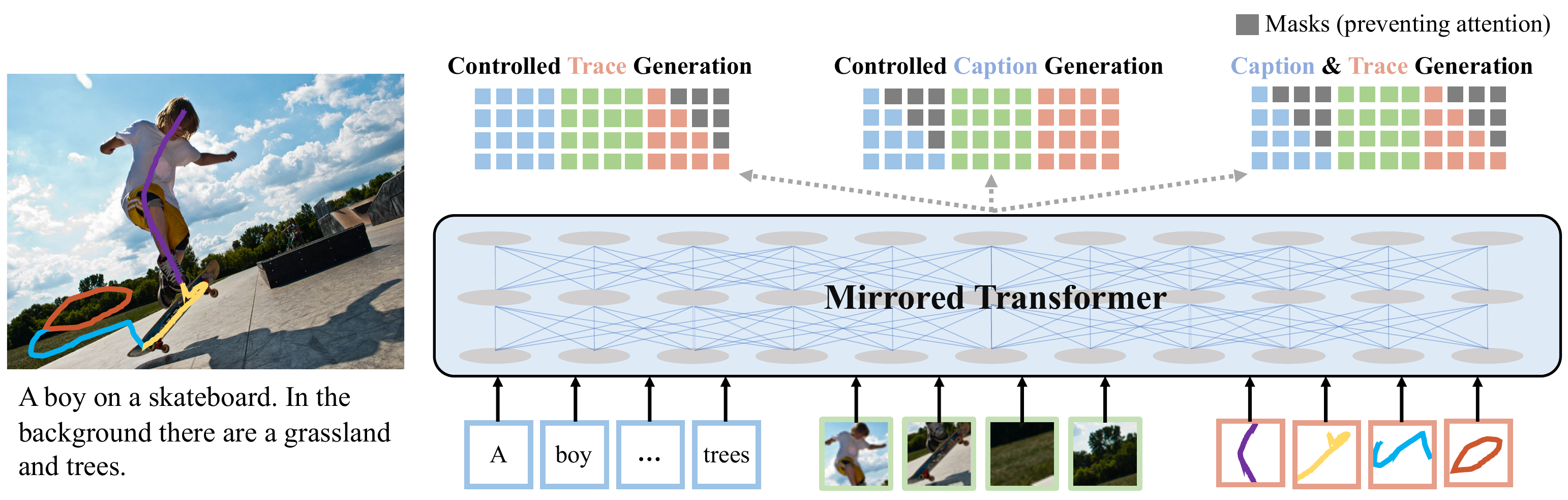} 
\caption{Overall architecture. Our proposed Mirrored Transformer (MITR) architecture effectively addresses the three tasks together by sharing most of the network modules. The structure is mirrored for processing the caption and trace. Depending on the task, we add a masking operation for the encoding/decoding of each module.}
\label{fig_overall_arc}
\end{figure*}


\subsection{Three Tasks on the Three Modalities}
\label{subsection:define_3tasks}
We first introduce how we encode the new trace modality.
The trace is a series of points with corresponding timestamps, each associated with a single word from the caption.
Instead of encoding each individual point, we convert the trace into a sequence of word-aligned bounding boxes, i.e., one box per word.
This encoding mitigates the local ``drawing" variation (by different annotators) within the same region, and thus more reliably allows the model to attend to the full spatial extent of a referred region.


In order to generate this dense word-to-box alignment from the provided trace points, we take the following steps: 
($i$) Split the trace into segments, with one segment per word (using the word-to-trace alignment from Localized Narratives).
($ii$) Generate one bounding box per trace segment, by taking the axis-aligned minimum bounding box of the convex hull of the mouse points. Then we introduce the three tasks:
\paragraph{Controlled Trace Generation}
Given an image $I$ and a caption describing this image $\mathbf{w} =\{ w_1, w_2, \cdots, w_N \}$, the model is required to generate a trace indicating the visual grounding corresponding to the caption, in the form of an ordered region sequence $\mathbf{r}=\{r_1, r_2, \cdots, r_T\}$.

\paragraph{Controlled Caption Generation} 
Given an image $I$ and a mouse trace provided by the user that is mapped to a sequence of regions, $\mathbf{r}=\{r_1, r_2, \cdots, r_T\}$, the model generates a caption $\mathbf{w} =\{ w_1, w_2, \cdots, w_N \}$ describing the image along this trace. 

\paragraph{Joint Caption and Visual Trace Generation}
We further propose a task which can be regarded as an extension of standard image captioning: given an image $I$, the model generates both caption $\mathbf{w} =\{ w_1, w_2, ..., w_N \}$ and its corresponding trace of ordered regions $\mathbf{r}=\{r_1, r_2 ..., r_T\}$ that matches the caption. 

\subsection{Mirrored Transformer for Three Modalities}
\label{subsection:mirrored_transformer}

\begin{figure}[!ht]
\centering
\includegraphics[width=0.40\textwidth]{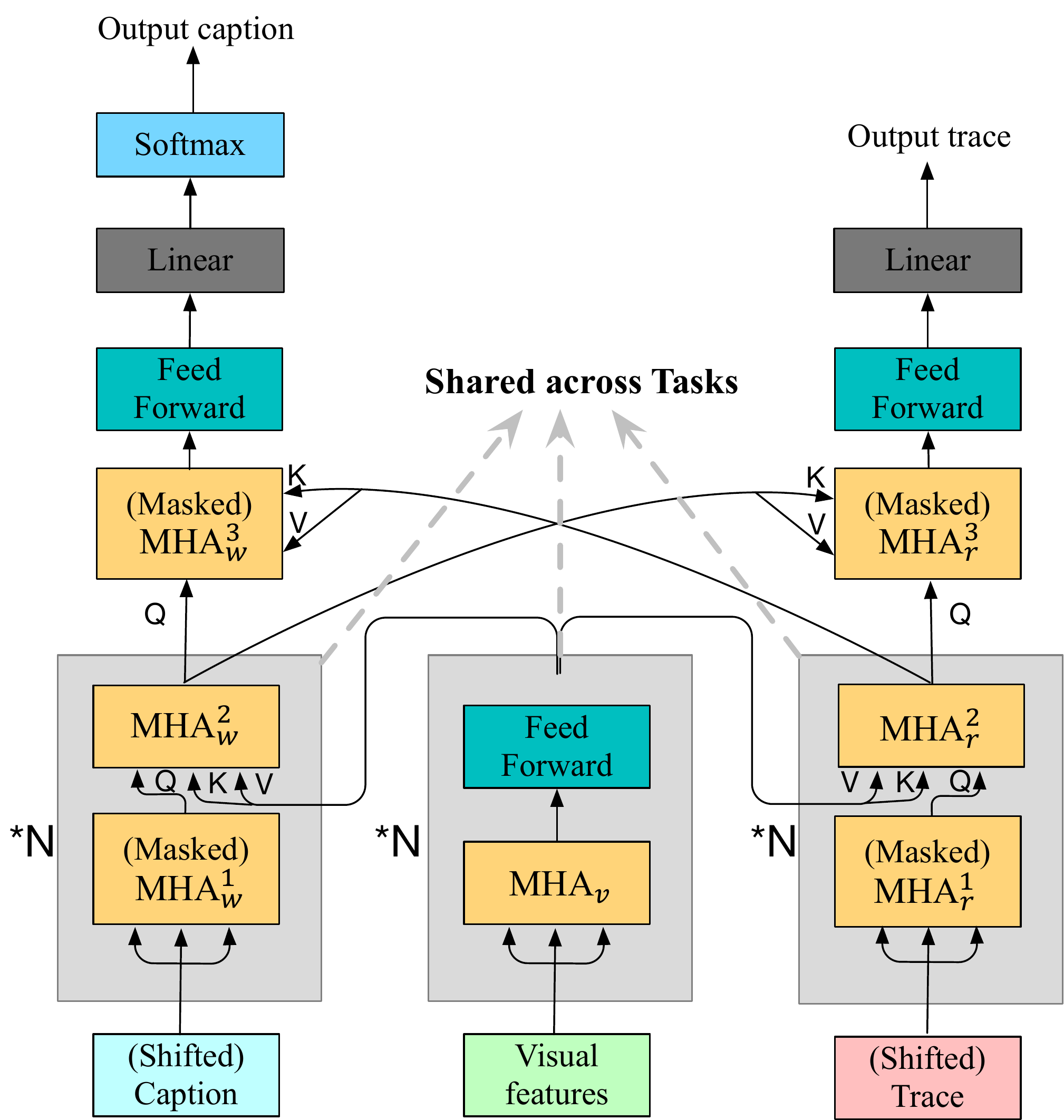} 
\caption{Mirrored Transformer (\modelName) architecture. MHA stands for MultiHead Attention.}
\label{fig_transformer_arc}
\end{figure}
Although the three tasks defined above are quite different, they operate on the same set of three modalities: image, caption, and trace. 
In this work, we propose a model that effectively addresses all three tasks together in a unified framework with shared parameters, rather than building three separate models.
Due to its symmetric structure, we name this model architecture ``\textbf{MI}rrored \textbf{T}ransforme\textbf{R}" (\modelName), as in Fig.~\ref{fig_overall_arc}.

\paragraph{Features}
The inputs to the model are subsets of: image features, text features, and trace features. 
For image features, we use pre-trained Faster R-CNN~\cite{anderson2018bottom} to compute the visual features of the detected regions.
For the text feature, we sum up the positional embeddings and the word embeddings, as in~\cite{vaswani2017attention}, where the position refers to the index of the word within the caption. 
For the trace feature, we sum up the positional embeddings and the input trace, which is projected into $d$ hidden dimensions. 
Specifically, we define the trace position as the index of the  bounding box that is aligned with the word in the corresponding caption.
We denote the input visual features, text features, and trace features as $x_v, x_w, x_r$, respectively. 

\paragraph{Model Architecture}
As in Fig.~\ref{fig_transformer_arc}, our model is composed of three modules (corresponding to three modalities): image encoder, caption encoder-decoder, and trace encoder-decoder. 
Each module consists of a transformer with self-attention.
Specifically, the image encoder, $h_v$, is defined as:
\begin{equation}
\label{eq_h_v}
    h_v = \text{FFN}(\text{MultiHead}_v(x_v, x_v, x_v)),
\end{equation}
where we follow \cite{vaswani2017attention} to define the feed-forward network (FFN) as two linear transformation layers with a ReLU activation in between, and the MultiHead as:
\begin{equation}\nonumber
    \text{MultiHead}(Q,K,V) = \text{Concat}(\text{head}_1, ..., \text{head}_c)W^O
\end{equation}
\begin{equation}\nonumber
    \text{head}_i = \text{Attention}(QW_i^Q, KW_i^K,VW_i^V),
\end{equation}
where the projections are parameter matrices. 
We refer readers to \cite{vaswani2017attention} for additional details of MultiHead attention.
Note that there is no masking operation in the MultiHead module from Eqn.~\eqref{eq_h_v}, since we allow the model to attend to all visual features when processing the caption and trace.

We then design a mirrored structure for the caption and trace modules, based on the observation that the two modalities are symmetric in the controlled caption generation and controlled trace generation tasks. 
The caption encoder-decoder, $h_w$, and trace encoder-decoder, $h_r$, are defined as:
\begin{equation}
\label{eq_h_w}\nonumber
    h_w = \text{MultiHead}_w^2(\text{MultiHead}_w^1(x_w,x_w,x_w), h_v, h_v)
\end{equation}
\begin{equation}
\label{eq_h_r}\nonumber
    h_r = \text{MultiHead}_r^2(\text{MultiHead}_r^1(x_r,x_r,x_r), h_v, h_v)
\end{equation}
Our caption and trace modules can switch roles between encoder and decoder seamlessly.
Inspired by~\cite{zhou2020unified}, this switching is implemented by a masking operation, where the encoder observes all inputs but the decoder only observes partial previous information.
This prevents the decoder from attending to future information. 
We implement a masking operation in either MultiHead$_w^1$ or MultiHead$_r^1$, depending on the specified task: 
\begin{itemize}
    \setlength\itemsep{0em}
    \item For controlled caption generation, the input caption is shifted right by one position, and MultiHead$_w^1$ applies masking to prevent leftward information flow. This ensures every position can only see its previous positions in the attention module. Note, the input trace is not shifted and MultiHead$_r^1$ does not have any masking. 
    \item For controlled trace generation, the input trace is shifted right by one position and MultiHead$_r^1$ applies masking, while the input caption is not shifted and MultiHead$_w^1$ does not perform the mask operation.
    \item For the joint caption and trace generation task, both the input caption and input trace are shifted right by one position, and both MultiHead$_w^1$ and MultiHead$_r^1$ perform mask operations.
\end{itemize}

Our model also supports multiple layers.
The module between $x_v,x_w,x_r$ and $h_v,h_w,h_r$ can be repeated $N$ times. 
Specifically, MultiHead$_v$ acts as the encoder, while MultiHead$^1_w$, MultiHead$^2_w$, MultiHead$^1_r$, and MultiHead$^2_r$ switch roles between encoder and decoder depending on what task is being performed. 
All of these modules are shared across different tasks.

Finally, once $h_w$ and $h_r$ have been computed, MultiHead$^3_w$ and MultiHead$^3_r$ are used to fuse the information from caption and trace modules. 
Note that in the joint caption and trace generation task, both MultiHead$^3_w$ and MultiHead$^3_r$ need to include a mask operation, while in the other two tasks, no mask operation is needed.

\subsection{Controlled Trace Generation: Distance Score}
\label{subsection:eval_metrics}
Given a ground truth trace of length $q$, represented as a sequence of $q$ bounding boxes, and a predicted trace of length $m$, we need a score that can measure the distance between these two traces. 
When $q=m$, the most straightforward way is to compute the $L1$ loss between pairs of bounding boxes (where the two bounding boxes at the same index in the sequence form a pair):
$D(\mathbf{r}^{gt}, \hat{\mathbf{r}}) = \frac{1}{q}\sum_{i=1}^{q} | r_i^{gt} - \hat{r_i} |$,
where $|r_i^{gt} - \hat{r_i} |$ is the mean $L1$ distance on the four coordinates of the $i$-th bounding box.

However, there are two main challenges. 
First, when $q\neq m$, we need to find the exact alignment between the two sets of bounding boxes. 
Second, even when $q=m$, we may not want to force the two sets to match in the given order because the dataset may contain examples where the local bounding box ordering is not semantically meaningful. 
\cite{chakraborty2017geometric} shows that if we treat each ``trajectory'' as a sequence of 
points on a Riemannian manifold,
a distance metric between two trajectories can be derived on a homogeneous space. 
Such an idea is useful for the case where the sample dimension is much larger than the number of samples. 
In our case, the sample dimension is small (4-D vector to represent each bounding box), so we choose to simply cast the evaluation task as a simple bipartite matching problem.
Note that standard bipartite matching is not a direct solution as it operates on two unordered sets of samples, ignoring the ordering within a trace. 
Instead, we propose to add local constraints to the bipartite matching so that the orderless matching can only happen within a local window. 
On the one hand, it provides a way to match two ordered sequences of bounding boxes; on the other hand, it allows local disorder, which is robust to the noise in the dataset annotation.

Consider two traces of lengths $q$ and $m$; without loss of generality, we assume that $q \leq m$. 
Let $C\in \mathbb{R}^{q\times m}$ be the cost matrix where $C_{ij}$ is the mean $L1$ distance between the four coordinates of the $i$-th box from the first trace and $j$-th box from the second trace, and let $X$ be the assignment matrix. 
We solve the following linear programming problem to get the distance between these two traces:
\begin{align*}
    \min_X \quad &\Tr (CX^T) \\
    s.t., \quad & X\mathbf{1}_m = \mathbf{1}_q,\quad  X^T\mathbf{1}_q \leq \mathbf{1}_m,\quad X\geq 0, \quad X_{i,j} = 0,\\
    &\forall i,j \quad s.t., \quad 0\leq i\leq q-1,\quad 0\leq j \leq m-1,\\ &j<\floor*{(i-k)\frac{m}{q}} \quad \text{or}  \quad  j \geq (i+1+k)\frac{m}{q},
\end{align*}
where $\mathbf{1}_m\in \mathbb{R}^{m}$ is all one vector and $k$ is the window size controlling the local range of disordered matching. 
For example, when $q=m$ and $k=1$, this allows one box from the first trace to match with the box at the same index from the second trace and also its left and right neighbors. 
After solving this linear programming problem, we use $\text{Tr}(CX^T) / q$ as the distance score between two traces. 
We call our proposed score the Local Bipartite Matching score (LBM). {In addition to using this as an evaluation measure, one 
could further incorporate this score into training to learn traces for a better matching score, by utilizing recently proposed differentiable linear programming solvers \cite{meng2020physarum,agrawal2019differentiable}. To keep the presentation succinct, we do not discuss these extensions in this paper.}

\subsection{Cycle Interaction of Trace and Caption}
Another interesting finding of our model architecture is that the controlled trace generation and controlled caption generation are dual problems in one framework, i.e., the output of one direction serves as the input of the other.
This inspires us to allow the two modules interact with each other.
First, we randomly permute the trace and feed it into the controlled caption generation module, generating the caption.
Then, we feed this generated caption\footnote{Gumbel-softmax~\cite{jang2016categorical} is applied to approximate the non-differentiable categorical sampling of words.} into our controlled trace generation model and enforce that the predicted trace be close to the originally permuted trace by adding a cycle loss. 
By doing so, we enrich the training set by adding more meaningful but unseen trace-caption pairs. 
As shown in Section~\ref{subsection:cycle_loss}, this further boosts the performance of both tasks.

We denote our mirrored transformer model as $f()$, the controlled trace generation task as $\hat{\mathbf{r}}=f(I, \mathbf{w})$, and the controlled caption generation task as $\hat{\mathbf{w}} = f(I, \mathbf{r})$. 
We enforce the cycle consistency via
\begin{equation}\label{eq:cycle_loss}\nonumber
    L_{\mathbf{\Tilde{r}}\to \mathbf{\hat{w}}\to \mathbf{\hat{r}}} =  \text{Dist}_{\mathbf{r}}(f(I, f(I,\mathbf{\Tilde{r}})), \mathbf{\Tilde{r}}),
\end{equation}
where $\text{Dist}_\mathbf{r}$ is the $L1$ loss between the predicted trace and the ground truth trace, and $\mathbf{\Tilde{r}}$ is the randomly manipulated trace. 
Specifically, we perform two types of manipulation: ($i$) randomly switch the trace within a mini-batch, and ($ii$) cut a trace into $S$ segments and randomly permute these segments to form a new trace.
We show that both manipulations are effective in improving the performance.

\subsection{Total Loss Function}
\label{subsec:total_loss}
The final loss function can be formulated as:
\begin{align}\label{eq:total_loss}
     L_{\text{total}} & = \lambda_1 L_{\text{[trace]}} + \lambda_2 L_{\text{[caption]}}\nonumber \\
     &+ \lambda_3 L_{\mathbf{\Tilde{r}}\to \mathbf{\hat{w}}\to \mathbf{\hat{r}}} + \lambda_4 L_{\text{[joint]}},
\end{align}
where $L_{[\text{trace}]}$ is the $L1$ loss between the predicted trace boxes and ground truth trace boxes for controlled trace generation, $L_{[\text{caption}]}$ is the cross-entropy loss of the caption for controlled caption generation, $L_{\mathbf{\Tilde{r}}\to \mathbf{\hat{w}}\to \mathbf{\hat{r}}}$ is the cycle loss, and $L_{\text{[joint]}}$ is the sum of the trace loss and the caption loss for the joint caption and trace generation task. 


\subsection{Bridging the Gap between Training and Testing}
Discrepancies between training and inference always exist in sequential prediction models~ \cite{bengio2015scheduled}. At training, the ground-truth input/output trace at each time step is provided, while at inference the unknown previous trace is replaced by a trace generated by the model itself.

In our proposed joint caption and trace generation task, such discrepancies are even more severe than in standard caption generation, as both the previous word and trace box are generated by the model and thus are connected.
The generated trace especially suffers from noise because, unlike the caption, the trace lacks syntax.
A single offset could cause the following trace boxes to quickly move to anywhere in the image.
To alleviate this problem, we propose a random replacement of the input trace boxes, where we replace a box with $[0,0,1,1,1]$ (corresponding to the whole image) with probability $p$. As shown in Table~\ref{tab:task3}, this approach improves the performance of joint caption and trace generation by a clear margin. 

%% file: text/4.experiments.tex
\section{Experiments}
\subsection{Dataset}

\begin{table}
    \centering
    \small
    \setlength{\tabcolsep}{4pt}
    \scalebox{0.95}{
    \begin{tabular}{l | c c c}
        \thickhline
         & \# images & \# captions & \# words/capt \\
        \hline
        COCO Loc. Narr. \cite{pont2019connecting} & 123,287 & 142,845 & 41.8 \\
        \hline
        Flickr30k Loc. Narr. \cite{pont2019connecting} & 31,783 & 32,578 & 57.1 \\
        \hline
        ADE20k Loc. Narr. \cite{pont2019connecting} & 22,210 & 22,529 & 43.0 \\
        \hline
        Open Images Loc. Narr. \cite{pont2019connecting} & 671,469 & 675,155 & 34.2 \\
        \hline
        \thickhline
    \end{tabular}
    }
    \caption{Localized Narratives built on top of COCO, Flickr30k, ADE20k, and Open Images.}
    \label{tab:datasets}
\end{table}

We conduct experiments on four datasets: COCO, Flickr30k, ADE20k, and Open Images, with annotations from two different frameworks: COCO Captions~\cite{chen2015microsoft} and Localized Narratives~\cite{pont2019connecting}, summarized in Table~\ref{tab:datasets}. We perform an ablation study on COCO and report the performance of our best performing model on the other three datasets in Localized Narratives and the downstream task introduced in Section~\ref{subsection:downstream_tasks} (evaluated on the COCO Captions annotations). We use the COCO2017 split for all experiments except Section~\ref{subsection:downstream_tasks}, where we follow \cite{cornia2019show} to use the split they provide. 

The annotations provided by Localized Narratives are challenging to work with for a number of reasons.
First, the human-generated trace segment $\mathbf{r}_i$ is a noisy visual representation of the mentioned object, due to imperfect voice-trace synchronization (e.g., if the annotator moves their mouse without speaking), errors in voice-word synchronization (from the automatic sequence-to-sequence alignment model in~\cite{pont2019connecting}), inconsistent drawing habits among annotators, and the different nature of mouse trace lines vs. bounding boxes. 
Second, not every word can be meaningfully grounded in the image, such as existentials (e.g., ``there are") and language referring to the observer (e.g., ``in this image, I can see ..."). By our estimate, such words account for at least 20\% of the words in the COCO validation captions from Localized Narratives. Traces for such words are less meaningful than the other groundable words.

\begin{table*}[!ht]
    \centering
    \small
    \setlength{\tabcolsep}{4pt}
    \resizebox{1.5\columnwidth}{!}{
    \begin{tabular}{l l | c c c c c c }
        \thickhline
        Method & Trained on & BLEU-1 & BLEU-4 & METEOR & ROUGE$_L$ & CIDEr & SPICE \\
        \hline 
        \cite{pont2019connecting} & Task2 & 0.522 & 0.246 & N/A & 0.483 & 1.065 & 0.365 \\
        \hline
        Baseline & Task2 & 0.563  & 0.255 & 0.240 & 0.453 & 0.997 &0.293 \ \\
        \modelName & Task2 & 0.577 & 0.257 & 0.245 & 0.456 & 1.213 & 0.293 \\
        \modelName & Task2 + Task1 & 0.586 & 0.272 & 0.252 & 0.470 & 1.329 & 0.307 \\
        \modelName & Task2 + Task1 + cycle$_s$ & 0.596 & 0.282 & 0.257 & 0.476 & 1.390 & 0.309 \\
        \modelName & Task2 + Task1 + cycle$_b$ & 0.598 & 0.286 & 0.258 & 0.479 & 1.407 & 0.313 \\
        \modelName(2 layer) & Task2 + Task1 + cycle$_b$ & \textbf{0.607} & \textbf{0.292} & \textbf{0.263} & \textbf{0.487} & \textbf{1.485} & \textbf{0.317} \\
        \thickhline
    \end{tabular}
    }
    \caption{Quantitative results for Task 2 (controlled caption generation) on COCO. cycle$_s$ and cycle$_b$ refer to two types of cycle loss defined in Sec~\ref{subsection:cycle_loss}. Results from \cite{pont2019connecting} are not directly comparable to ours due to differences mentioned in Sec~\ref{subsec:individual_tasks}.}
    \label{tab:task2}
\end{table*}

\begin{table*}[!ht]
    \centering
    \small
    \setlength{\tabcolsep}{4pt}
    \resizebox{1.95\columnwidth}{!}{
    \begin{tabular}{l l | c c c c c c c c}
        \thickhline
        Method & Trained on & BLEU-1 & BLEU-4 & METEOR & ROUGE$_L$ & CIDEr & SPICE & LBM(k=0) & LBM(k=1) \\
        \hline 
        Baseline & Task2 & 0.355 & 0.087 & 0.155 & 0.307 & \textbf{0.310} & 0.210 & N/A & N/A \\
        \modelName & Task3 & 0.387 & 0.118 & 0.168 & 0.316 & 0.170 & 0.194 & 0.387 & 0.369 \ \\
        \modelName & Task3 + random mask & 0.395 & \textbf{0.128} & \textbf{0.184} & \textbf{0.328} & 0.219 & \textbf{0.223} & 0.308 & 0.292 \\
        \modelName & Task3 + Task1 + Task2 + random mask & \textbf{0.417} & 0.125 & 0.178 & 0.323 & 0.216 & 0.213 & \textbf{0.283} & \textbf{0.267} \\
        \thickhline
    \end{tabular}
    }
    \caption{Quantitative results for Task 3 (joint caption and trace generation) on COCO.}
    \label{tab:task3}
\end{table*}

\begin{table*}[!ht]
    \centering
    \small
    \setlength{\tabcolsep}{4pt}
    \resizebox{1.95\columnwidth}{!}{
    \begin{tabular}{l l l| c c c c c c c c}
        \thickhline
        Dataset & Method & Trained on & BLEU-1 & BLEU-4 & METEOR & ROUGE$_L$ & CIDEr & SPICE & LBM(k=0) & LBM(k=1) \\
        \hline 
        Flickr30k & Baseline & Task1& N/A & N/A & N/A & N/A & N/A & N/A & 0.253 & 0.249\\
        Flickr30k & Baseline & Task2 & 0.620 & 0.345 & 0.286 & 0.524 & 1.763 & 0.341 & N/A & N/A\\
        Flickr30k & \modelName & Task1 + Task2 + cycle$_b$ &   \textbf{0.644} & \textbf{0.374} & \textbf{0.300} & \textbf{0.547} & \textbf{2.014} & 0.365 & 0.195 & 0.188\\
        Flickr30k & \modelName(2 layer) & Task1 + Task2 + cycle$_b$ & 0.640 & 0.366 & 0.297 & 0.543 & 2.000 & \textbf{0.374} & \textbf{0.183} & \textbf{0.176}  \\
        \hline
        ADE20k & Baseline & Task1 & N/A & N/A & N/A & N/A & N/A & N/A & 0.251 & 0.247\\
        ADE20k & Baseline & Task2 & 0.565 & 0.278 & 0.259 & 0.475 & 1.288 & 0.341 & N/A & N/A\\
        ADE20k & \modelName & Task1 + Task2 + cycle$_b$ & \textbf{0.580} & \textbf{0.297} & \textbf{0.269} & \textbf{0.490} & \textbf{1.463} & 0.354  & \textbf{0.177} & \textbf{0.168}\\
        ADE20k & \modelName(2 layer)  & Task1 + Task2 + cycle$_b$ & 0.575 & 0.292 & 0.267 & 0.490 & 1.437 & \textbf{0.358} & 0.191 & 0.185 \\
        \hline
        Open Images & Baseline & Task1 & N/A & N/A & N/A & N/A & N/A & N/A & 0.212 & 0.209 \\
        Open Images & Baseline & Task2 & 0.560 & 0.273 & 0.261 & 0.503 & 1.467 & 0.361 & N/A & N/A \\
        Open Images & \modelName & Task1 + Task2 + cycle$_b$ & 0.573 & 0.292 & 0.271 & 0.520 & 1.584 & 0.372 & \textbf{0.180} & \textbf{0.171} \\
        Open Images & \modelName(2 layer)  & Task1 + Task2 + cycle$_b$ & \textbf{0.584} & \textbf{0.308} & \textbf{0.276} & \textbf{0.531} & \textbf{1.887} & \textbf{0.381} & 0.211 & 0.206  \\
        
        \thickhline
    \end{tabular}
    }
    \caption{Quantitative results for Task 1 and Task2 on Localized Narratives of Flickr30k, ADE20k and Open Images.}
    \label{tab:task1_2_three_datasets}
\end{table*}

        

\subsection{Experimental Setting}
\label{subsec:exp_setting}
We use our mirrored transformer (\modelName) defined in Section~\ref{subsection:mirrored_transformer} with $N=1$ (as shown in Fig.~\ref{fig_transformer_arc}). The hidden size of attention layers is $512$ and that of the feed-forward layers is $2048$. We train the network with batch size $30$ using the Adam optimizer \cite{kingma2014adam}. The initial learning rate is $5\mathrm{e}{-4}$, which decays every $3$ epochs with decay rate $0.8$, for a total of $30$ epochs. We use the same training setup for all experiments reported in this paper. The random masking rate for joint caption and trace prediction is $p=0.5$. In the following, we denote controlled trace generation as Task$1$, controlled caption generation as Task$2$, and joint caption and trace generation as Task$3$. 

\paragraph{Controlled Trace Generation}
In this task, we represent the trace as an ordered sequence of bounding boxes, and the model predicts one bounding box for each word of the input caption, as described in Section~\ref{subsection:define_3tasks}. Given a ground truth trace $\mathbf{r}^{gt}=\{r_1^{gt}, r_2^{gt}, ..., r_T^{gt}\}$ and a predicted trace $\mathbf{\hat{r}}=\{\hat{r}_1, \hat{r}_2, ..., \hat{r}_T\}$ for the same image,  we compute the local bipartite matching (LBM) score proposed in Section~\ref{subsection:eval_metrics} for $k=0$ and $k=1$. 

\paragraph{Controlled Caption Generation}
Given an image and a trace, the model predicts the caption corresponding to the trace. When evaluating the quality of generated captions, we report the following widely adopted metrics: BLEU-$1$, BLEU-$4$ \cite{BLEU}, METEOR \cite{METEOR}, ROGUE \cite{ROUGE}, CIDEr \cite{CIDEr}, SPICE \cite{SPICE}.
We use beam search with size $5$. 

\paragraph{Joint Caption and Trace Generation} In this task, the model is given only an image as input and outputs both caption and trace simultaneously. The model produces outputs iteratively, generating one word and one corresponding bounding box at each time step. At test time, we end the generation when the caption generation branch outputs the \texttt{END} token. In this process, since the model itself controls the length of the output, the length of the predicted trace $\mathbf{\hat{r}}$ may differ from the length of the ground truth trace $\mathbf{r}^{gt}$. The max length for generating caption and trace is set to be $100$. We report our LBM metric for both $k=0$ and $k=1$. 

\paragraph{Baselines}
For controlled trace generation, we construct the baseline by using a standard one-layer encoder-decoder transformer architecture as defined in \cite{vaswani2017attention} and feed both visual features and captions to the encoder. Similarly, for controlled caption generation, we use the same architecture as a baseline and feed both visual features and traces to the encoder. For joint caption and trace generation, we construct the baseline by also using the same architecture, but only using visual features as input, and we train it on the caption generation task.

\subsection{Results on Individual Tasks}
\label{subsec:individual_tasks}
\begin{figure*}[!ht]
\centering
\includegraphics[width=0.88\textwidth]{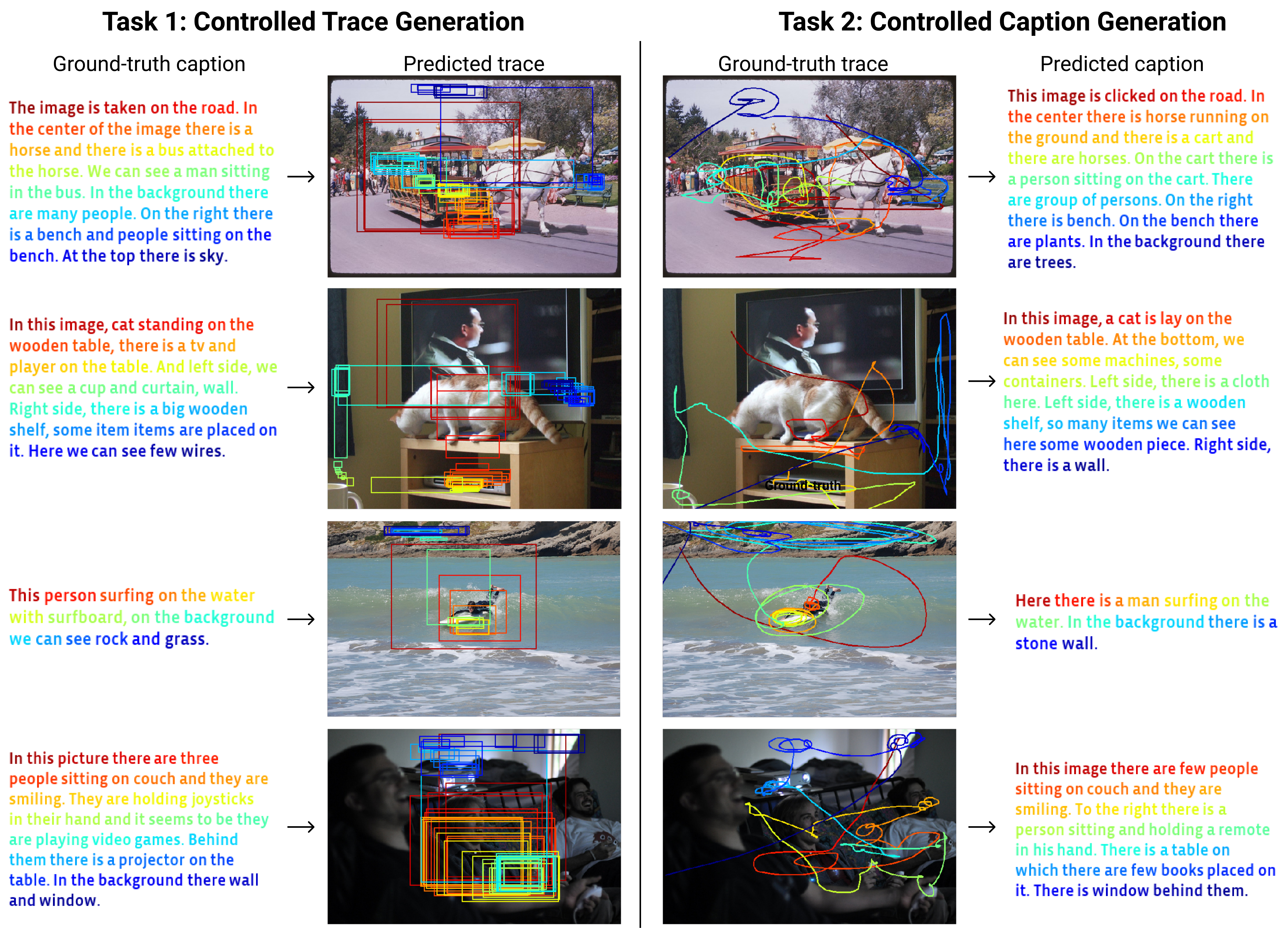} 
\caption{Qualitative results on Tasks 1 and 2 on COCO. (See supplement for visualizations for other three datasets)}
\label{fig:qual_results_tasks_1_2_coco}
\end{figure*}

\paragraph{Controlled Trace Generation} In Fig.~\ref{fig:qual_results_tasks_1_2_coco} (left), we demonstrate the qualitative results of the controlled trace generation: we can see that the trace closely follows the ground truth trace and also semantically corresponds well to the input caption. Table \ref{tab:task1} shows the quantitative results. 
From the table, we see that our proposed MITR outperforms the baseline method constructed in Section~\ref{subsec:exp_setting} (standard transformer with one encoder and one decoder \cite{vaswani2017attention}). 

\begin{table}[!ht]
    \centering
    \small
    \setlength{\tabcolsep}{4pt}
    \resizebox{1.0\columnwidth}{!}{
    \begin{tabular}{l l | c c }
        \thickhline
        Method & Trained on & LBM (k=0) & LBM (k=1) \\
        \hline 
        Baseline & Task1 & 0.208 & 0.204 \\
        \modelName & Task1 & 0.171  & 0.159 \\
        \modelName & Task1 + Task2 & 0.169 & 0.157 \\
        \modelName & Task1 + Task2 + cycle$_s$ & 0.165 & 0.156 \\
        \modelName & Task1 + Task2 + cycle$_b$ & 0.166 & 0.155 \\
        \modelName (2 layer) & Task1 + Task2 + cycle$_b$ & \textbf{0.163} & \textbf{0.154} \\
        \thickhline
    \end{tabular}
    }
    \caption{Quantitative results for Task 1 (controlled trace generation). cycle$_s$ and cycle$_b$ refer to the two cycle losses defined in Sec~\ref{subsection:cycle_loss}. Note: smaller values of LBM are better.}
    \label{tab:task1}
\end{table}

\begin{figure*}[!ht]
\centering
\includegraphics[width=0.88\textwidth]{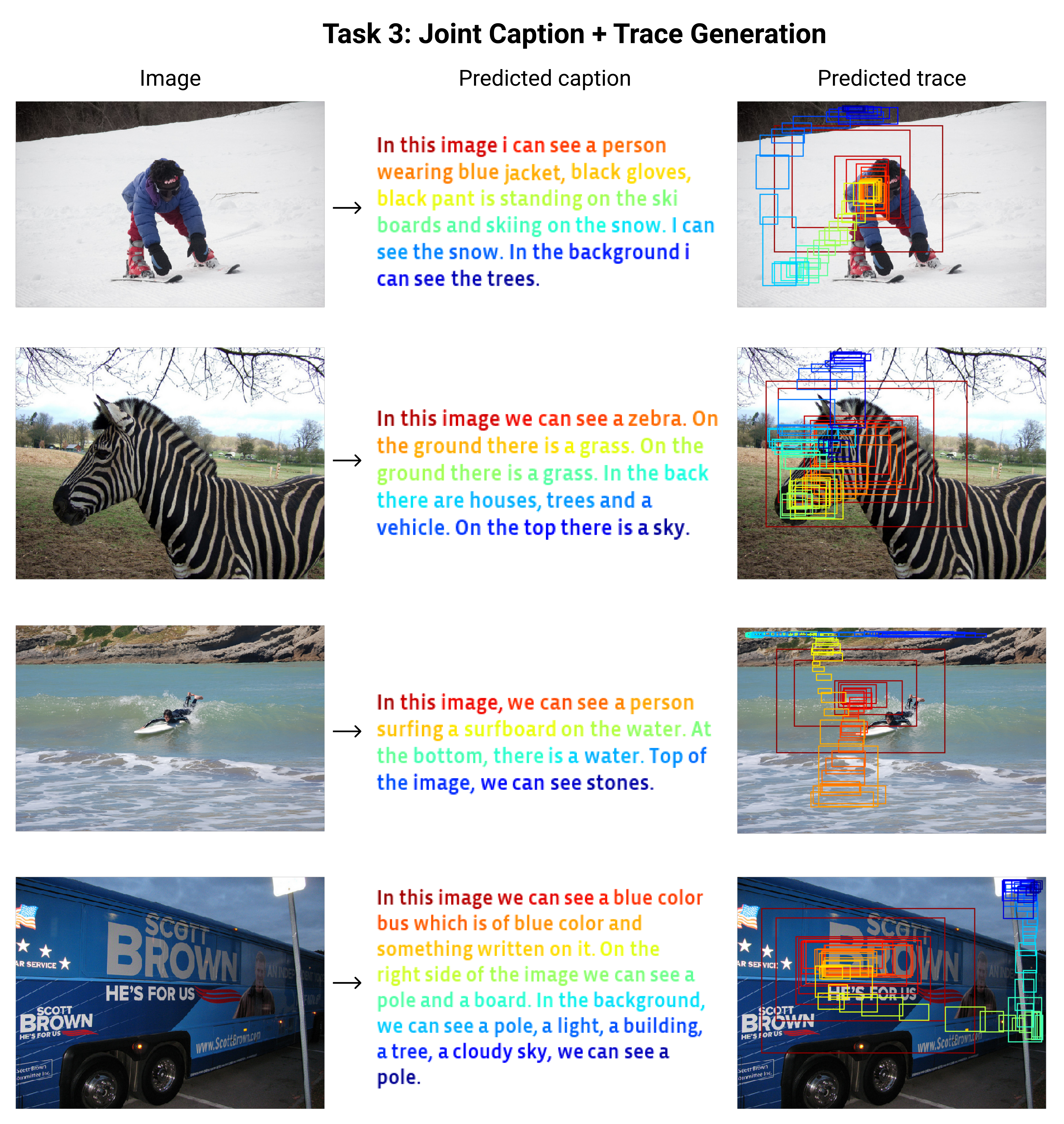} 
\caption{Qualitative results on Task 3 on COCO.}
\label{fig:qual_results_tasks_3_coco}
\end{figure*}


\paragraph{Controlled Caption Generation}  
We show our quantitative results in Table~\ref{tab:task2} and qualitative results on the right side in Fig.~\ref{fig:qual_results_tasks_1_2_coco}. 
Our baseline model differs from the one in \cite{pont2019connecting} in several places: 
we use a one-layer encoder-decoder transformer while \cite{pont2019connecting} uses two layers;
in addition, we process the trace by cutting the trace by word while \cite{pont2019connecting} cuts the trace by a fixed time interval. 
Thus, results from \cite{pont2019connecting} are not directly comparable to ours. 
We mention the performance of \cite{pont2019connecting} in Table \ref{tab:task2} for easy reference.


\paragraph{Joint Caption and Trace Generation} The quantitative results for this task are in Table~\ref{tab:task3}. We can see that by modeling the trace at the same time as the caption, the performance of caption generation improves by a large margin over the baseline, which only models the caption. In addition, our proposed random masking technique further improves the performance of Task 3 on caption generation by over $1\%$ absolute improvement on all metrics, and on trace generation by nearly $20\%$ relative improvement. The qualitative results are shown in Fig.~\ref{fig:qual_results_tasks_3_coco}. Without human annotated attention traces to guide the caption generation, sometimes the same objects or descriptions are repeated multiple times in a single caption. 
This suggests that future developments must keep an account of all the objects referenced to avoid repetition.

\subsection{Joint Training Results}
We demonstrate that, by performing joint training, our model can boost the performance of each individual task while using approximately one half of the parameters and training compute cost, compared with training one separate model for each individual task. The quantitative results are in Tables~\ref{tab:task1} and~\ref{tab:task2}. 
Further, we can see from Table \ref{tab:task3} that the joint training of Task 1 and Task 2 can also help Task 3.

\subsection{Cycle Loss Results}
\label{subsection:cycle_loss}
We show that, by enforcing cycle consistency, both controlled trace generation and controlled caption generation are further improved when joint training is used. The quantitative results are in Tables~\ref{tab:task1} and \ref{tab:task2}. We use cycle$_s$ to represent cycle loss where a single trace is cut into segments and then randomly permuted before forming a new trace, and cycle$_b$ to represent cycle loss where the trace is permuted along the batch dimension within a mini-batch. Adding cycle$_b$ achieves over $1\%$ absolute improvement on BLEU-1 and BLEU-4 compared with our joint training result and over $3\%$ absolute improvement from our baseline model.

\subsection{Results on Flickr30k, ADE20k, Open Images} 
\label{subsection:three_datasets}
We also report the performance of our best performing model and the baseline model on another three datasets (Flickr30k, AED20k, Open Images), where Localized Narratives are also collected~\cite{pont2019connecting}. 
The results are given in Table \ref{tab:task1_2_three_datasets}. 
As shown, our method achieves consistent improvement over the baseline methods on all datasets.

\begin{table}[!h]
    \centering
    \small
    \setlength{\tabcolsep}{4pt}
    \scalebox{0.95}{
    \begin{tabular}{l | c c c c c c }
        \thickhline
          & B-1 & B-4 & M & R & C & S \\
        \hline 
        Ours \emph{w/o} pretrain & 0.463  & 0.182 & 0.219 & 0.466 & 1.746 & 0.363 \ \\
        Ours \emph{w/} pretrain  & \textbf{0.474} & \textbf{0.189} & \textbf{0.225} & \textbf{0.475} & \textbf{1.819} & \textbf{0.370} \\
        \thickhline
    \end{tabular}
    }
    \caption{Downstream task on guided caption generation.}
    \label{tab:task_downstream}
\end{table}

\subsection{Downstream Task}
\label{subsection:downstream_tasks}
We further investigate the benefit of our joint training framework. 
By pre-training using our joint training framework on Localized Narratives \cite{pont2019connecting} and fine-tuning on a guided caption generation task \cite{cornia2019show} on COCO Captions~\cite{chen2015microsoft}, we are able to get better results than directly training on COCO Captions. 
In this experiment, we follow~\cite{cornia2019show} to use the COCO split provided by~\cite{karpathy2015deep}.

The task is defined as: given an image $I$ and a sequence of ordered bounding boxes $\mathbf{r}=\{r_1, r_2,\cdots, r_T\}$ as guidance, the model generates a caption $\mathbf{w} =\{ w_1, w_2, \cdots, w_N \}$. This task is similar to our controlled caption generation task (Task 2), but we do not assume any correspondence between the boxes and words for both training and testing. 
Note that \cite{cornia2019show} considers a slightly different setting, where the dense correspondences between boxes and words are given during training but not at testing.
Thus a special gate function was proposed to automatically attend the words to the boxes during test time.  
See the supplementary material for more details. 
The results are in Table~\ref{tab:task_downstream}, where pre-training offers a clear gain.


%% file: text/5.conclusion.tex
\section{Conclusion}
We presented a unified framework for modeling vision, language, and human attention traces.
Our work is built on top of the Localized Narratives framework and motivated by the need for longform image captions and dense visual grounding.
We proposed a Mirrored Transformer model architecture that was jointly trained on three vision-and-language tasks.
We demonstrated the effectiveness of our approach through detailed experiments on four datasets.

%% file: Localized Narratives/supp.tex
\section{Visualizations on Tasks 1 and 2 on ADE20k, Flickr30k and Open Image}
\begin{figure*}[!h]
\centering
\includegraphics[width=0.88\textwidth]{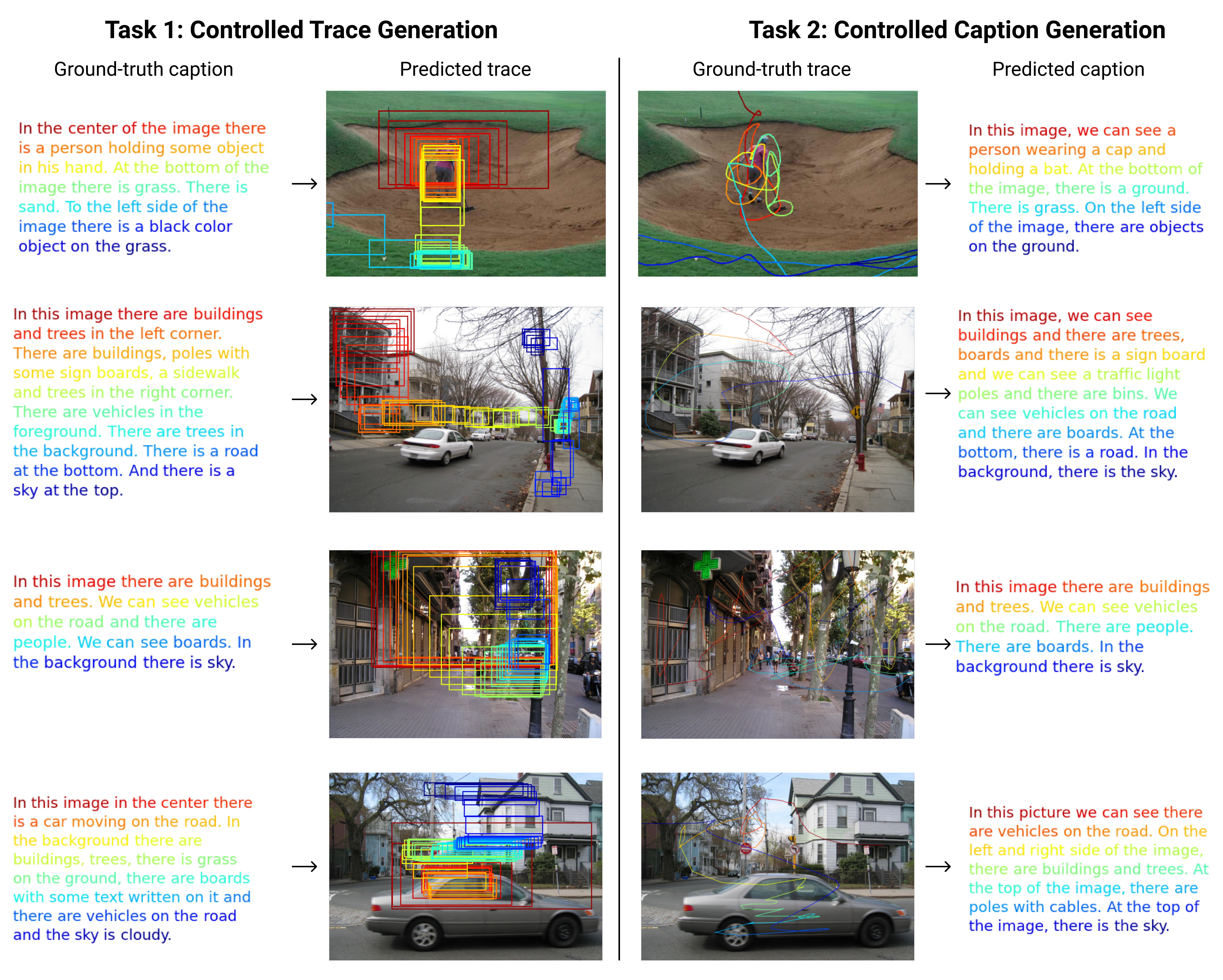} 
\caption{Qualitative results on Tasks 1 and 2 on ADE20k.}
\label{fig:qual_results_tasks_1_2_ade20k}
\end{figure*}

\begin{figure*}[!h]
\centering
\includegraphics[width=0.88\textwidth]{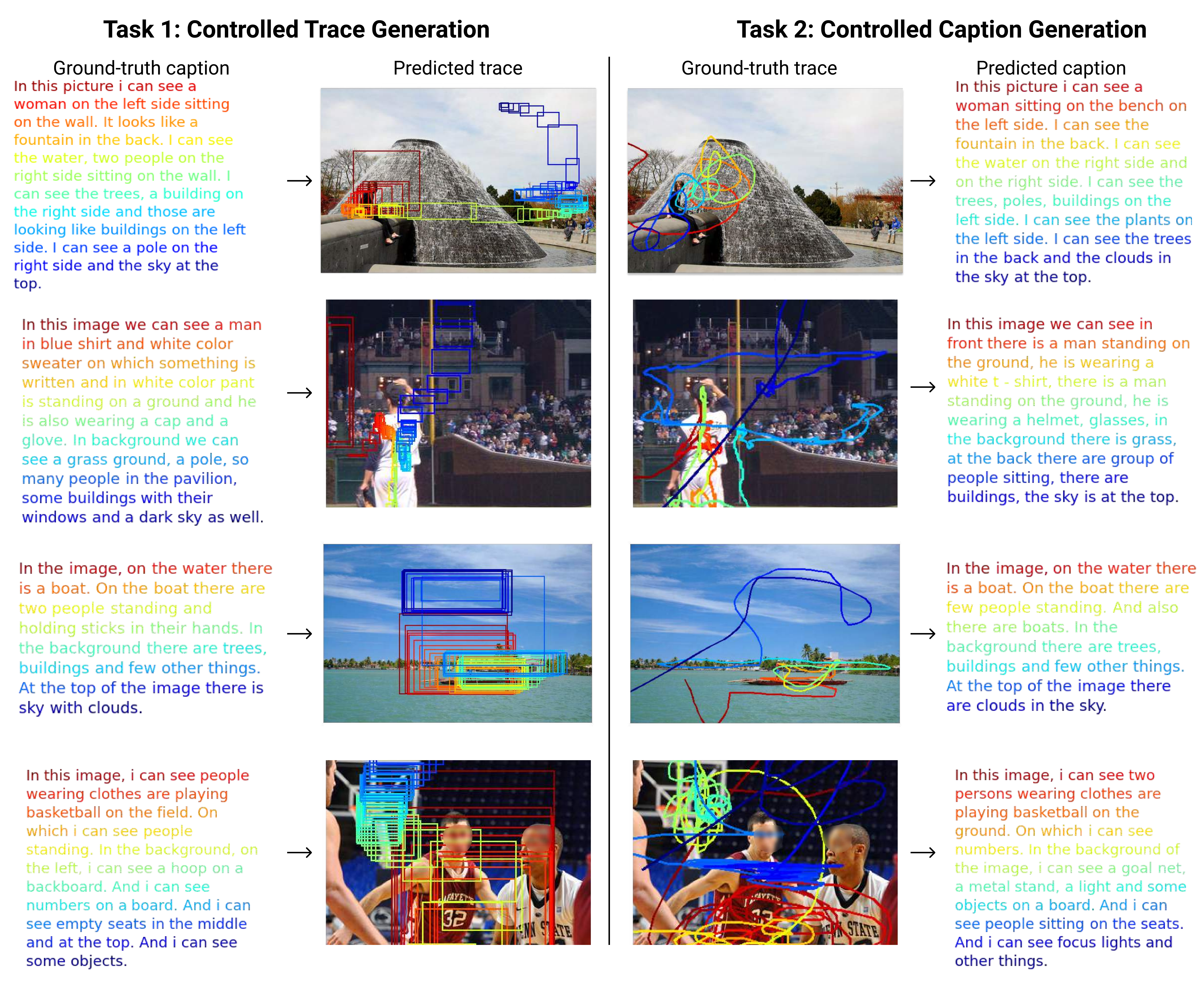} 
\caption{Qualitative results on Tasks 1 and 2 on Flickr30k.}
\label{fig:qual_results_tasks_1_2_flk30k}
\end{figure*}

\begin{figure*}[!h]
\centering
\includegraphics[width=0.88\textwidth]{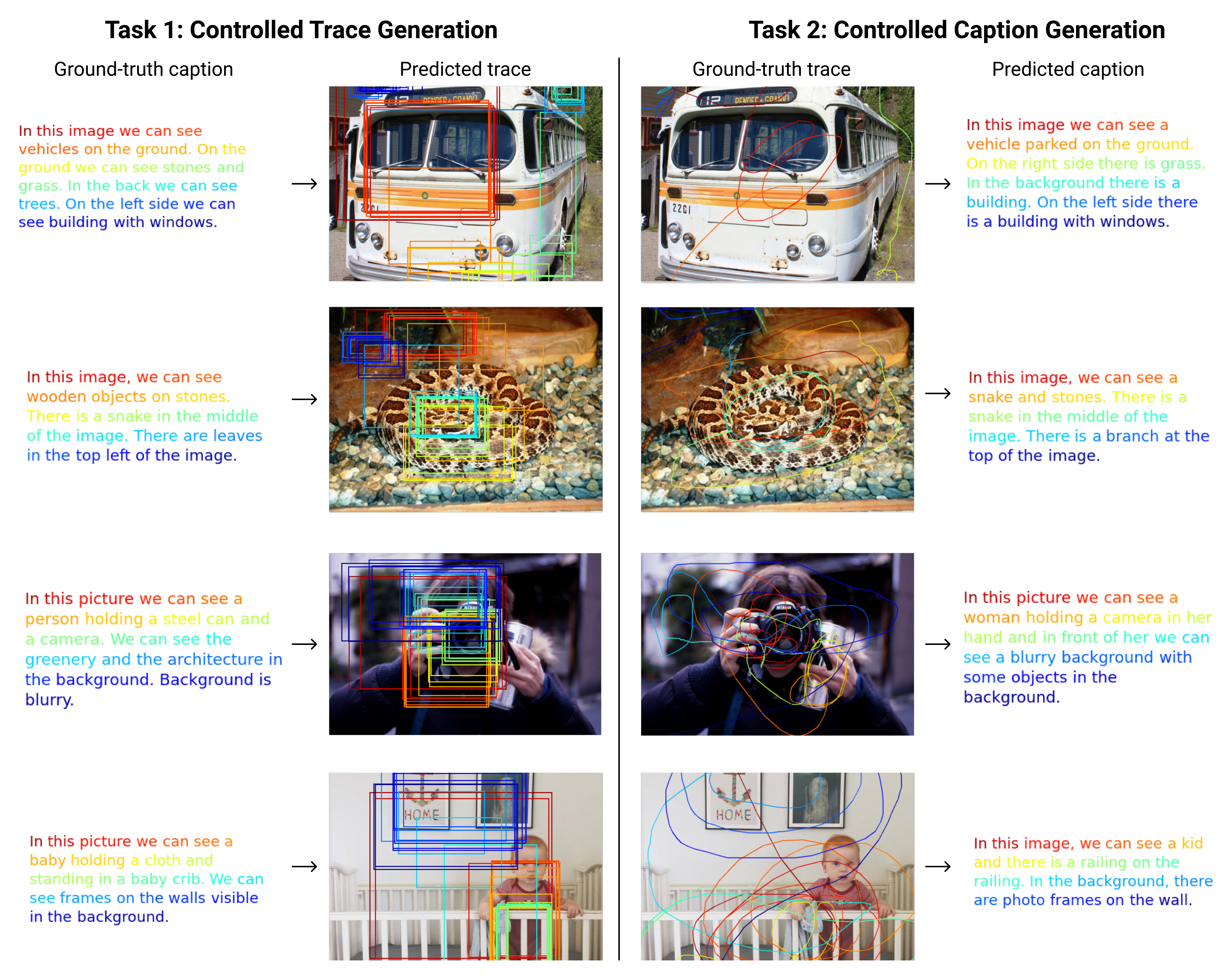} 
\caption{Qualitative results on Tasks 1 and 2 on Open Image.}
\label{fig:qual_results_tasks_1_2_openimg}
\end{figure*}

See Fig. \ref{fig:qual_results_tasks_1_2_ade20k} \ref{fig:qual_results_tasks_1_2_flk30k} \ref{fig:qual_results_tasks_1_2_openimg}.

\section{Implementation Details of Downstream Task on COCO Captions}
In the key boxes guided caption generation task defined by \cite{cornia2019show}, at training time, we are provided a dense correspondence between caption and bounding boxes that associates every word in the caption with a bounding box detected by Faster-RCNN. 
If a word has no associated bounding box, \cite{cornia2019show} uses the average feature from all detected bounding boxes in this image as its corresponding visual representation.
At inference time, the only input is the image with several ordered bounding boxes given by the user -- no knowledge of which bounding box corresponds to which word(s) in the generated caption is assumed. 
\cite{cornia2019show} addressed this problem by proposing a specialized gate function to learn how to attend each word to the given boxes at inference time. 

In contrast, our experiment proceeds under a slightly different setting: 
we are given the same information for both training and inference, i.e., only a sequence of bounding boxes for an image dense correspondence between boxes and words are not provided. We have the same inference setting as ~\cite{cornia2019show} but 
it is a more challenging training setting than~\cite{cornia2019show}, because we are provide sparser alignment of boxes and words during training. 

Also note that this new task is different from the controlled caption generation task in our main paper.
In controlled caption generation, the input is a meaningful smooth trace describing overall image content that often includes the relationship between objects and the background in the image, but our downstream task only utilizes several key box-object pairs.

To adapt such input into the same form as the setting where we deal with localized narratives dataset, we simply concatenate the given bounding boxes into a sequence, and pad $[0,0,1,1,1]$ if the length is less than what is needed (e.g., the length of sentence).
In this way, the input becomes the same form (although the contained information is not exactly the same) that we used on the Localized Narratives experiments. 
Principally the specialized gate function proposed by \cite{cornia2019show} can also be added in our network, but we simply choose to use the same input format as our experiments on Localized Narratives because our goal here is to demonstrate the benefit of pre-training instead of trying to reproduce the setting in \cite{cornia2019show}. This results in a slightly different setting than that defined in \cite{cornia2019show}. As shown in the main paper, our pre-training on Localized Narratives brings clear gain under this new setting.

\section{Layer Choice for Mirrored Transformer}
We demonstrate the influence of the number of layers in our proposed mirrored transformer by varying the number of layers in the model trained with Task$1$ + Task$2$ + cycle$_b$ (cycle loss by permuting the trace within a mini batch).
The results are shown in Table \ref{tab:layers}, which shows that two layers lead to better performance compared to one layer on both controlled trace generation (Task1) and controlled caption generation (Task2), while using three layers does not further improve results.

\begin{table*}[!ht]
    \centering
    \begin{tabular}{c | c c c c c c c c}
        \thickhline
         $\#$layers & BLEU-1 & BLEU-4 & METEOR & ROUGE$_L$ & CIDEr & SPICE & LBM ($k=0$) & LBM ($k=1$) \\
        \hline 
        1 & 0.598 & 0.286 & 0.258 & 0.479 & 1.407 & 0.313 & 0.166 & 0.155 \\
       2 & \textbf{0.607} & \textbf{0.292} & \textbf{0.263} & \textbf{0.487} & \textbf{1.485} & \textbf{0.317} & \textbf{0.163} & \textbf{0.154} \\
        3 & 0.604 & 0.289 & 0.261 & 0.485 & 1.444 & 0.314 & 0.197 & 0.191 \\
        \thickhline
    \end{tabular}
    \caption{Influence of the number of layers. The model was trained on Task$1$ + Task$2$ + cycle$_b$, and evaluated on Task$1$ (LBM metric) and Task$2$ (other metrics in this table) respectively. Note: smaller values of LBM are better. The evaluation is performed on COCO2017 Validation set.}
    \label{tab:layers}
\end{table*}

\section{Influence of $\lambda$ in Joint Training}
We use the joint training of Task1 and Task2 as an example to show the influence of different $\lambda$ values (defined in Eq. (3) in our main paper). 
The results are shown in Table \ref{tab:lambda}. We can see that the performance of Task2 (controlled caption generation) remains relatively stable across different $\lambda$, while the performance of Task1 (controlled trace generation) improves when Task$1$ has a larger weight compared with Task$2$. 
In the experiments of the main paper, all values of $\lambda$ are chosen from $\{1.0, 0.5, 0.3, 0.1, 0.0\}$ according to specific experiment settings and the performance on a subset of $5000$ images from COCO2017 Training set (which we use to tune the hyperparameters).

\begin{table*}[!ht]
    \centering
    \begin{tabular}{l | c c c c c c c c}
        \thickhline
         $\lambda_2$ & BLEU-1 & BLEU-4 & METEOR & ROUGE$_L$ & CIDEr & SPICE & LBM ($k=0$) & LBM ($k=1$) \\
        \hline 
        1.0 & 0.589 & 0.272 & 0.254 & 0.472 & 1.346 & 0.306 & 0.187 & 0.178 \\
        0.5 & \textbf{0.595} & 0.278 & \textbf{0.256} & \textbf{0.476} & 1.368 & \textbf{0.312} & 0.179 & 0.169 \\
        0.3 & 0.586 & 0.272 & 0.252 & 0.470 & 1.329 & 0.307 & 0.169 & 0.157 \\
        0.1 & \textbf{0.595} & \textbf{0.279} & \textbf{0.256} & 0.474 & \textbf{1.375} & 0.310 & \textbf{0.161} & \textbf{0.150} \\
      
        \thickhline
    \end{tabular}
    \caption{Results on different $\lambda_2$ when $\lambda_1=1$ for joint training of Task$1$ and Task$2$ ($\lambda$ defined in eq. (3) in our main paper). The model was trained on Task$1$ + Task$2$, and evaluated on Task$1$ (LBM metric) and Task$2$ (other metrics in this table) respectively. Note: smaller values of LBM are better. The evaluation is done on COCO2017 Validation set.}
    \label{tab:lambda}
\end{table*}

\section{More Qualitative Results}

\begin{figure*}[!ht]
\centering
\includegraphics[width=0.8\textwidth]{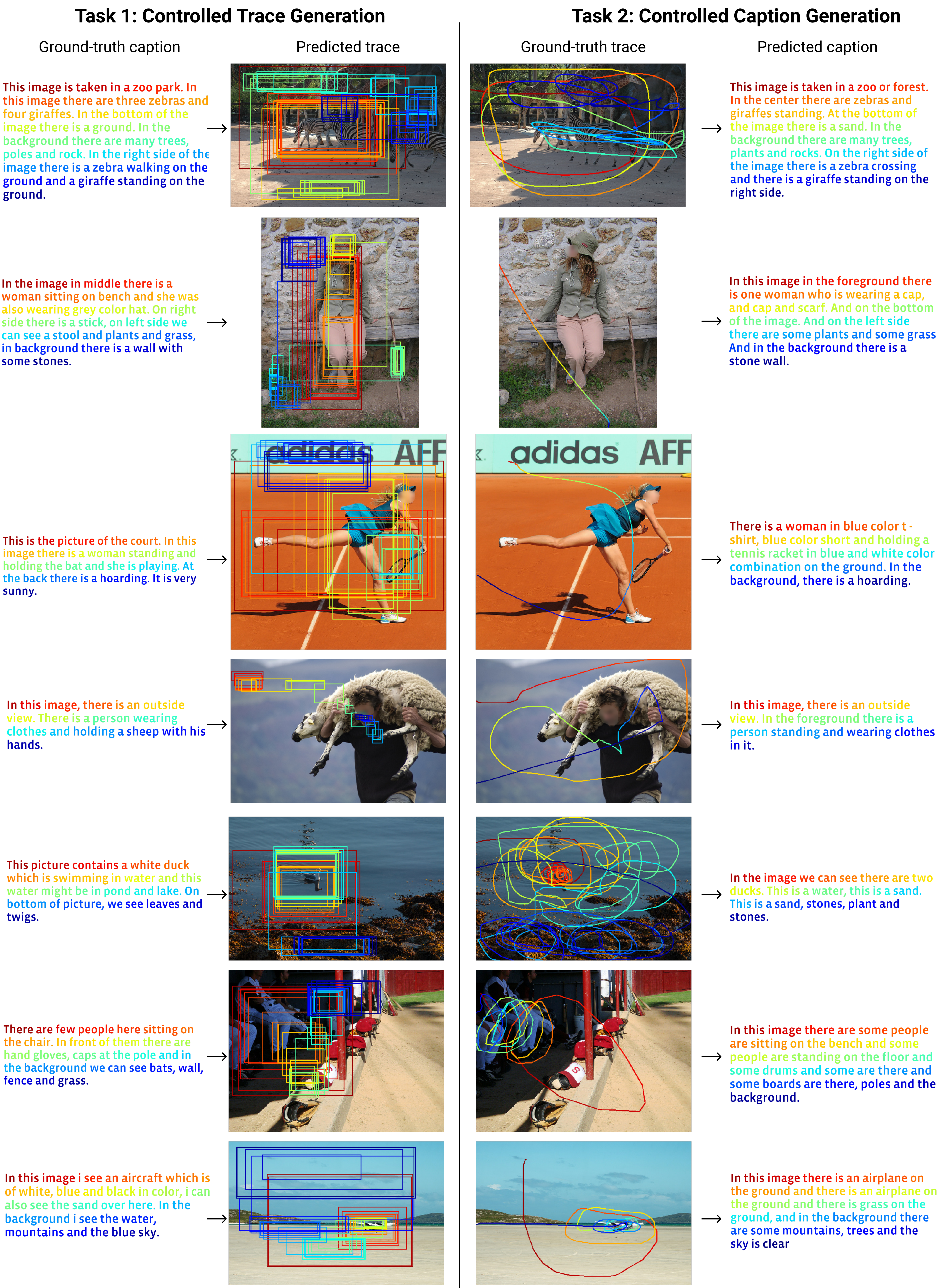}
\caption{Qualitative results and selected failure cases on Tasks 1 and 2. The model was trained on Task$1$ + Task$2$ + cycle$_b$.}
\label{fig:qual_results_tasks_1_and_2_supplement}
\end{figure*}

This section shows more qualitative results and analysis, for both success and failure cases of the model. 
In each sub-section, the failure cases are ordered in descending order of subtlety (i.e., most obvious to most subtle). 

\subsection{Controlled Trace Generation}
\label{subsection:qual_errors_task1}
Below, we describe some successful instances and common failure cases of the model on the \emph{controlled trace generation} task. Examples are all in Fig.~\ref{fig:qual_results_tasks_1_and_2_supplement}, on the \textbf{left} column.

\subsubsection{Successes}
\paragraph{Correct object localization and spatial extent.} The model successfully localizes the referred to objects and identifies their full spatial extents. For example, see row 1 (the animals and trees), row 2 (the woman, her hat, the stick), and row 3 (the tennis court, the woman, the racket, and the ``adidas" text).

\paragraph{Recognition of directionality.} The model attends to direction words in the input caption, such as ``right" and ``left." For example, in row 1, the caption specifies that there is a zebra and a giraffe in the ``\textbf{right} side of the image" and the predicted trace correctly localizes these animals, even though there are other zebras and giraffes in the image that are described earlier in the caption. In row 2, where the caption begins ``In the image in middle ...", the model quickly narrows in on the middle of the image (the red bounding boxes), rather than localizing the entire image. In row 3, the model correctly localizes the ``\textbf{back}" of the image when the caption refers to this area.

\paragraph{Adapting to errors in the input caption.} The model is able to adapt to errors in the input caption, such as spelling errors and incorrect object classifications. For example, in row 3, the model successfully localizes the ``adidas" text when the caption reads ``hoarding" (perhaps the annotator meant ``heading"), and also localizes the tennis racket, even though it is referred to as a ``bat."

\subsubsection{Failure cases}
\paragraph{False negatives in the trace.} A ground-truth object or concept is not localized by the predicted trace. For example, see row 6. The predicted trace does not include the baseball bats and grass. This could be due to two reasons: (1) the model recognizes the bats and fence as relevant, but incorrectly localizes them, or (2) the model is focusing on larger and more visually dominant false positive objects in the image (such as the red columns in this image), and neglects the bats and fence from the trace.

\paragraph{Incorrect object spatial extent.}  A predicted region has the correct localization (e.g., the bounding box is positioned in the center of the referred-to object), but its spatial extent does not cover the full object. For example, see row 4: the man and the sheep are correctly localized, but their predicted spatial extents are too small. Another example is row 7: the predicted box for ``sand" has good precision, in that it only localizes sand, but it does not cover the full spatial extent of the sand.

\paragraph{Poor region differentiation.}  Predicted regions referring to different objects/regions are correctly localized and have reasonable spatial extents, but these regions are not specific to each object. For example, see row 5. The duck and the water are correctly localized, but there is no way to differentiate these regions (since they all cover the same area around the duck). Ideally, the model would predict a tight box around the duck and a much larger box covering the entire water region.

\subsection{Controlled Caption Generation}
\label{subsection:qual_errors_task2}
In this section, we describe some successful instances and common failure cases of the model on the \emph{controlled caption generation} task. Examples are all in Fig.~\ref{fig:qual_results_tasks_1_and_2_supplement}, on the \textbf{right} column.

\subsubsection{Successes}
\paragraph{All major image components are described.} The model successfully describes all the objects and ``stuff" in the image, as compared to the ground-truth caption. See rows 1-3 for examples.

\paragraph{Caption is grammatical and tells a story.} The predicted caption uses proper grammar, introduces the image (e.g., ``In this image, there is ..."), and moves around the image describing different regions and objects. See rows 1-3 for examples.

\paragraph{Caption includes directionality.} The model uses directions to refer to specific regions of the image it is describing. For example, see row 1 (the zebras and giraffes on the ``\textbf{center} and ``\textbf{right side}" of the image) and row 2 (the woman in the ``\textbf{foreground}", the plants and grass on the ``\textbf{left side}", and the stone wall in the ``\textbf{background}".

\paragraph{Adapting to poor input traces.} The model can output a rich caption, even given a poor input trace where the annotator drew a trace that is uncorrelated with the ground-truth caption. See rows 2 and 3 for examples.

\subsubsection{Failure cases}

\paragraph{False negative objects in the caption.} Here, the caption omits mentioning a visually significant object or region that is specified by the human-provided trace. For example, see row 1 (the predicted caption neglects to mention the sheep) or row 7 (the model misses the sand).

\paragraph{False positive objects in the caption.} Here, the caption hallucinates objects that are not present in the image. For example, see row 6: the model incorrectly describes the image as containing ``drums."

\paragraph{Incorrect object counts.} In this failure case, the model predicts the wrong number of objects that are present in the image. For example, see row 5: the model incorrectly predicts that there are ``two ducks," rather than just one.

\paragraph{Object repetition.} Here, the model correctly identifies an object in the image (that only has one instance), but mentions it multiple times. For example, see row 7: the model mentions the airplane twice (``there is an airplane on the ground and there is an airplane on the ground"), even though the image contains only instance of ``airplane."

\paragraph{Grammar errors.} This is a fairly common error, where the contents of the caption are correct, but the model uses incorrect grammar. For example, see row 5 (``this is a water, this is a sand"), and row 6 (``and some there are and some boards are there"). In many cases, the ground-truth captions have incorrect grammar (for example, see Fig.~\ref{fig:qual_results_tasks_1_and_2_supplement}, left side, row 5), which could cause the model to learn and internalize these errors.

\subsection{Joint Caption And Trace Generation}
\label{subsection:qual_errors_task3}
\begin{figure}[!ht]
\centering
\includegraphics[width=0.5\textwidth]{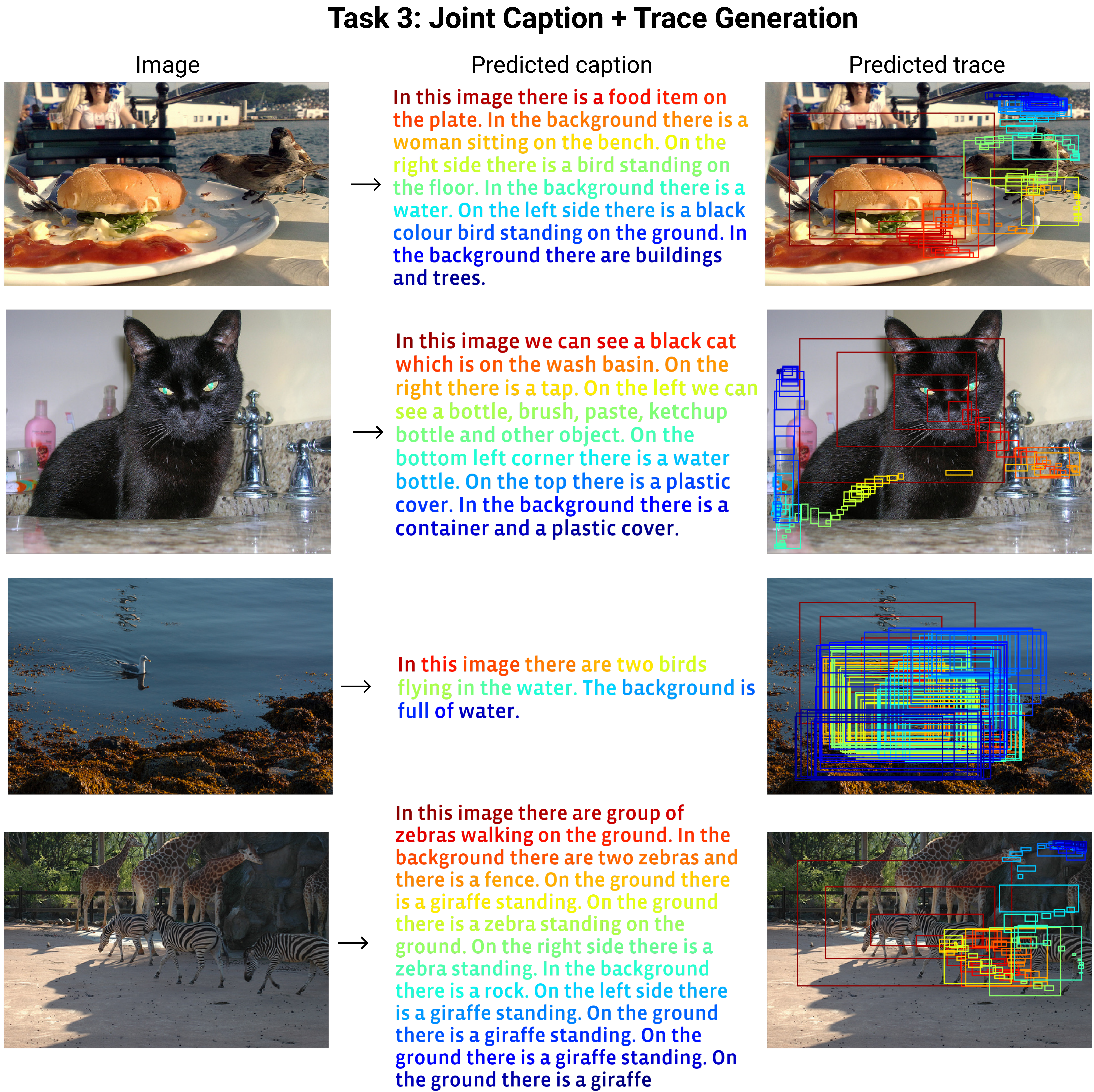}
\caption{Qualitative results and selected failure cases on Task 3. The model was trained on Task$3$ + random mask.}
\label{fig:qual_results_tasks_3_supplement}
\end{figure}

In this section, we describe some successful instances and common failure cases of the model on the \emph{joint caption and trace generation} task. The model may also experience the issues described in Sections~\ref{subsection:qual_errors_task1} and~\ref{subsection:qual_errors_task2}, but we focus on the errors specific to Task 3. Examples are all in Fig.~\ref{fig:qual_results_tasks_3_supplement}. 

\subsubsection{Successes}
Successful examples follow all the qualities of Task 1 and 2 (correct object localizations and spatial extents in the predicted trace, and precise, descriptive, and comprehensive predicted captions). They also have good alignment between the boxes in the predicted trace and the words in the caption. See rows 1 and 2 for examples.

\subsubsection{Failure cases}
\paragraph{Unaligned caption and trace.} In this failure case, the model predicts a much longer trace than what is reasonable for the caption. For example, see row 3: the predicted trace has length 100, while the caption only has 18 words.

\paragraph{Caption cuts off due to maximum length constraint.} In this case, the model is forced to stop predicting words because it hits the maximum caption length requirement (in this paper, this value is $100$). This error usually happens in images that contain many distinct objects, because it is challenging for the model to group objects for conciseness. See row 4 for an example: the caption ends in the middle of a sentence (``On the ground there is a giraffe"). One solution would be to enforce the maximum caption length \emph{and} require the caption to terminate with a full sentence (i.e., ending with a period), rather than allowing sentence fragments.

